\documentclass[11pt]{article}
\usepackage{eacl2017}
\usepackage{times}
\usepackage{url}
\usepackage{latexsym}

\usepackage{subfig}
\usepackage{graphicx}
\usepackage{amsmath}
\usepackage{multirow}
\usepackage{tabularx}

\eaclfinalcopy % Uncomment this line for the final submission
\def\eaclpaperid{55} %  Enter the acl Paper ID here

\title{Using Word Embedding for Cross-Language Plagiarism Detection}

\author{
  J\'{e}r\'{e}my Ferrero \\
  Compilatio \\
  276 rue du Mont Blanc \\
  74540 Saint-F\'{e}lix, France \\
  LIG-GETALP \\
  Univ. Grenoble Alpes, France \\
  {\tt jeremy.ferrero@imag.fr} \\
  \And
  Fr\'{e}d\'{e}ric Agn\`{e}s \\
  Compilatio \\
  276 rue du Mont Blanc \\
  74540 Saint-F\'{e}lix, France \\
  {\tt frederic@compilatio.net} \\
  \AND
  Laurent Besacier \\
  LIG-GETALP \\
  Univ. Grenoble Alpes, France \\
  {\tt laurent.besacier@imag.fr} \\
  \And
  Didier Schwab \\
  LIG-GETALP \\
  Univ. Grenoble Alpes, France \\
  {\tt didier.schwab@imag.fr} \\
}

\date{}

\begin{document}
\maketitle
\begin{abstract}
This paper proposes to use distributed representation of words (word embeddings) in cross-language textual similarity detection.
The main contributions of this paper are the following: (a) we introduce new cross-language similarity detection methods based on distributed representation of words; (b) we combine the different methods proposed to verify their complementarity and finally obtain an overall $F_{1}$~score of 89.15\% for English-French similarity detection at chunk level (88.5\% at sentence level) on a very challenging corpus.
\end{abstract}

\section{Introduction}
\label{intro}

Plagiarism is a very significant problem nowadays, specifically in higher education institutions.
In monolingual context, this problem is rather well treated by several recent researches \cite{plagiarism-overview}.
Nevertheless, the expansion of the Internet, which facilitates access to documents throughout the world and to increasingly efficient (freely available) machine translation tools, helps to spread \emph{cross-language plagiarism}.
Cross-language plagiarism means plagiarism by translation, \textit{i.e.} a text has been plagiarized while being translated (manually or automatically).
The challenge in detecting this kind of plagiarism is that the suspicious document is no longer in the same language of its source.
We investigate how distributed representations of words can help to propose new cross-lingual similarity measures, helpful for plagiarism detection.
We use word embeddings \cite{word2vec} that have shown promising performances for all kinds of NLP tasks, as shown in \newcite{eval-biling1}, \newcite{eval-biling2} and \newcite{eval-we}, for instance.
\newline\newline
\textbf{Contributions.}
The main contributions of this paper are the following:
\vspace{-0.3em}
\begin{itemize}
\setlength\itemsep{-1.3em}
\item we augment some state-of-the-art methods with the use of word embeddings instead of lexical resources; \\
\item we introduce a syntax weighting in distributed representations of sentences, and prove its usefulness for textual similarity detection; \\
\item we combine our methods to verify their complementarity and finally obtain an overall $F_{1}$~score of 89.15\% for English-French similarity detection at chunk level (88.5\% at sentence level) on a very challenging corpus (mix of Wikipedia, conference papers, product reviews, Europarl and JRC) while the best method alone hardly reaches $F_{1}$~score higher than 50\%.
\end{itemize}

\section{Evaluation Conditions}
\label{evaluation}

\subsection{Dataset}
\label{dataset}

The reference dataset used during our study is the new dataset recently introduced by \newcite{dataset-lrec}\footnote{\url{https://github.com/FerreroJeremy/Cross-Language-Dataset}}.
The dataset was specially designed for a rigorous evaluation of cross-language textual similarity detection.

More precisely, the characteristics of the dataset are the following:
\vspace{-0.3em}
\begin{itemize}
\setlength\itemsep{-0.3em}
\item it is multilingual: it contains French, English and Spanish texts;
\item it proposes cross-language alignment information at different granularities: document level, sentence level and chunk level;
\item it is based on both parallel and comparable corpora (mix of Wikipedia, conference papers, product reviews, Europarl and JRC);
\item it contains both human and machine translated texts;
\item it contains different percentages of named entities;
\item part of it has been obfuscated (to make the cross-language similarity detection more complicated) while the rest remains without noise;
\item the documents were written and translated by multiple types of authors (from average to professionals) and cover various fields.
\end{itemize}

In this paper, we only use the French and English sub-corpora.

\subsection{Overview of State-of-the-Art Methods}
\label{baselines}

%Textual similarity detection methods are not exactly methods to detect plagiarism.
Plagiarism is a statement that someone copied text deliberately without attribution, while these methods only detect textual similarities.
%There is no way of knowing why texts are similar and thus to assimilate these similarities to plagiarism.
However, textual similarity detection can be used to detect plagiarism.
\newline\newline
The aim of cross-language textual similarity detection is to estimate if two textual units in different languages express the same or not.
We quickly review below the state-of-the-art methods used in this paper, for more details, see \newcite{dataset-lrec}.

\textit{Cross-Language Character \mbox{N-Gram} (CL-C$n$G)} is based on \newcite{mcnamee2004} model.
We use the \newcite{potthast2011} implementation which compares two textual units under their \mbox{$3$-grams} vectors representation.

\textit{Cross-Language Conceptual Thesaurus-based Similarity (CL-CTS)} \cite{pataki2012} aims to measure the semantic similarity using abstract concepts from words in textual units.
In our implementation, these concepts are given by a linked lexical resource called \mbox{\textit{DBNary}} \cite{dbnary}.

\textit{Cross-Language Alignment-based Similarity Analysis (CL-ASA)} aims to determinate how a textual unit is potentially the translation of another textual unit using bilingual unigram dictionary which contains translations pairs (and their probabilities) extracted from a parallel corpus (\newcite{barron2008}, \newcite{pinto2009}).

\textit{Cross-Language Explicit Semantic Analysis (CL-ESA)} is based on the explicit semantic analysis model \cite{gabrilovich2007}, which represents the meaning of a document by a vector based on concepts derived from \mbox{Wikipedia}.
It was reused by \newcite{potthast2008} in the context of cross-language document retrieval.

\textit{Translation + Monolingual Analysis (T+MA)} consists in translating the two units into the same language, in order to operate a monolingual comparison between them \cite{barron2012}.
We use the \newcite{muhr2010} approach using \mbox{\textit{DBNary}} \cite{dbnary}, followed by monolingual matching based on bags of words.

\subsection{Evaluation Protocol}
\label{protocol}

We apply the same evaluation protocol as in \newcite{dataset-lrec}'s paper.
We build a distance matrix of size~$N$\,x\,$M$, with~$M$~=~1,000 and~$N$~=~$|S|$ where~$S$ is the evaluated sub-corpus.
Each textual unit of~$S$ is compared to itself (to its corresponding unit in the target language, since this is cross-lingual similarity detection) and to~$M$-1 other units randomly selected from~$S$.
The same unit may be selected several times.
Then, a matching score for each comparison performed is obtained, leading to the distance matrix.
Thresholding on the matrix is applied to find the threshold giving the best~$F_{1}$~score.
The~$F_{1}$~score is the harmonic mean of precision and recall.
Precision is defined as the proportion of relevant matches (similar cross-language units) retrieved among all the matches retrieved.
Recall is the proportion of relevant matches retrieved among all the relevant matches to retrieve.
Each method is applied on each \mbox{EN-FR} sub-corpus for chunk and sentence granularities.
For each configuration (\textit{i.e.} a particular method applied on a particular sub-corpus considering a particular granularity), 10~folds are carried out by changing the~$M$ selected units.

\section{Proposed Methods}
\label{methods}

The main idea of word embeddings is that their representation is obtained according to the context (the words around it).
The words are projected on a continuous space and those with similar context should be close in this multi-dimensional space.
A similarity between two word vectors can be measured by cosine similarity.
So using word-embeddings for plagiarism detection is appealing since they can be used to calculate similarity between sentences in the same or in two different languages (they capture intrinsically synonymy and morphological closeness).
We use the \mbox{\textit{MultiVec}} \cite{multivec} toolkit for computing and managing the continuous representations of the texts.
It includes word2vec \cite{word2vec}, paragraph vector \cite{paragraph-vector} and bilingual distributed representations \cite{bivec} features.
The corpus used to build the vectors is the News Commentary\footnote{\url{http://www.statmt.org/wmt14/translation-task.html}} parallel corpus.
For training our embeddings, we use CBOW model with a vector size of 100, a window size of 5, a negative sampling parameter of 5, and an alpha of 0.02.

\subsection{Improving Textual Similarity Using Word Embeddings (\emph{\mbox{CL-CTS-WE}} and \emph{\mbox{CL-WES}})}

We introduce two new methods. First, we propose to replace the lexical resource used in \emph{\mbox{CL-CTS}} (\textit{i.e.} \mbox{\textit{DBNary}}) by distributed representation of words. We call this new implementation \emph{\mbox{CL-CTS-WE}}.
More precisely, \emph{\mbox{CL-CTS-WE}} uses the top~10 closest words in the embeddings model to build the BOW of a word.
Secondly, we implement a more straightforward method (\emph{\mbox{CL-WES}}), which performs a direct comparison between two sentences in different languages, through the use of word embeddings.
It consists in a cosine similarity on distributed representations of the sentences, which are the summation of the embeddings vectors of each word of the sentences.

Let~$U$ a textual unit, the~$n$ words of the unit are represented by~$u_{i}$ as:
\begin{equation}
U = \{u_{1}, u_{2}, u_{3}, ..., u_{n}\}
\end{equation}

If~$U_{x}$ and~$U_{y}$ are two textual units in two different languages, \emph{\mbox{CL-WES}} builds their (bilingual) common representation vectors~$V_{x}$ and~$V_{y}$ and applies a cosine similarity between them.

A distributed representation~$V$ of a textual unit~$U$ is calculated as follows:
\begin{equation}
V = \sum_{i=1}^{n}(vector(u_{i}))
\end{equation}
where~$u_{i}$ is the~$i^{th}$ word of the textual unit and $vector$ is the function which gives the word embedding vector of a word.
This feature is available in \mbox{\textit{MultiVec}}\footnote{\url{https://github.com/eske/multivec}} \cite{multivec}.

\subsection{Cross-Language Word Embedding-based Syntax Similarity (CL-WESS)}

Our next innovation is the improvement of \emph{\mbox{CL-WES}} by introducing a \textit{syntax flavour} in it.
Let~$U$ a textual unit, the~$n$ words of the unit are represented by~$u_{i}$ as expressed in the formula (1).
First, we syntactically tag $U$ with a part-of-speech tagger (\mbox{\textit{TreeTagger}} \cite{schmid1994}) and we normalize the tags with Universal Tagset of \newcite{universal-tagset}.
Then, we assign a weight to each type of tag: this weight will be used to compute the final vector representation of the unit.
Finally, we optimize the weights with the help of \emph{\mbox{Condor}} \cite{condor}.
\emph{\mbox{Condor}} applies a Newton\textquoteright s method with a trust region algorithm to determinate the weights that optimize the~$F_{1}$~score.
We use the first two folds of each sub-corpus to determinate the optimal weights. 

The formula of the syntactic aggregation is:
\begin{equation}
V = \sum_{i=1}^{n}(weight(pos(u_{i})).vector(u_{i}))
\end{equation}
where $u_{i}$ is the $i^{th}$ word of the textual unit, $pos$ is the function which gives the universal part-of-speech tag of a word, $weight$ is the function which gives the weight of a part-of-speech, $vector$ is the function which gives the word embedding vector of a word and $.$ is the scalar product.

If~$U_{x}$ and~$U_{y}$ are two textual units in two different languages, we build their representation vectors~$V_{x}$ and~$V_{y}$ following the formula (3) instead of (2), and apply a cosine similarity between them.
We call this method \emph{\mbox{CL-WESS}} and we have implemented it in \mbox{\textit{MultiVec}} \cite{multivec}.

It is important to note that, contrarily to what is done in other tasks such as neural parsing \cite{chenM14}, we did not use POS information as an additional vector input because we considered it would be more useful to use it to weight the contribution of each word to the sentence representation, according to its morpho-syntactic category.

\begin{table*}[ht]
 \begin{center}
\begin{small}
\begin{tabular}{|l|l|l|l|l|l|l|}
	  \hline
	  \multicolumn{7}{|c|}{\bf Chunk level} \\
      \hline
      \bf Methods & \bf Wikipedia (\%) & \bf TALN (\%) & \bf JRC (\%) & \bf APR (\%) & \bf Europarl (\%) & \bf Overall (\%) \\
      \hline
      CL-C3G & 63.04 {\scriptsize $\pm$ 0.867} & 40.80 {\scriptsize $\pm$ 0.542} & 36.80 {\scriptsize $\pm$ 0.842} & 80.69 {\scriptsize $\pm$ 0.525} & 53.26 {\scriptsize $\pm$ 0.639} & 50.76 {\scriptsize $\pm$ 0.684} \\
      CL-CTS & 58.05 {\scriptsize $\pm$ 0.563} & 33.66 {\scriptsize $\pm$ 0.411} & 30.15 {\scriptsize $\pm$ 0.799} & 67.88 {\scriptsize $\pm$ 0.959} & 45.31 {\scriptsize $\pm$ 0.612} & 42.84 {\scriptsize $\pm$ 0.682} \\
      CL-ASA & 23.70 {\scriptsize $\pm$ 0.617} & 23.24 {\scriptsize $\pm$ 0.433} & 33.06 {\scriptsize $\pm$ 1.007} & 26.34 {\scriptsize $\pm$ 1.329} & 55.45 {\scriptsize $\pm$ 0.748} & 47.32 {\scriptsize $\pm$ 0.852} \\
      CL-ESA & 64.86 {\scriptsize $\pm$ 0.741} & 23.73 {\scriptsize $\pm$ 0.675} & 13.91 {\scriptsize $\pm$ 0.890} & 23.01 {\scriptsize $\pm$ 0.834} & 13.98 {\scriptsize $\pm$ 0.583} & 14.81 {\scriptsize $\pm$ 0.681} \\
      T+MA & 58.26 {\scriptsize $\pm$ 0.832} & 38.90 {\scriptsize $\pm$ 0.525} & 28.81 {\scriptsize $\pm$ 0.565} & 73.25 {\scriptsize $\pm$ 0.660} & 36.60 {\scriptsize $\pm$ 1.277} & 37.12 {\scriptsize $\pm$ 1.043} \\
      \hline
      CL-CTS-WE & 58.00 {\scriptsize $\pm$ 1.679} & 38.04 {\scriptsize $\pm$ 2.072} & 31.73 {\scriptsize $\pm$ 0.875} & 73.13 {\scriptsize $\pm$ 2.185} & 49.91 {\scriptsize $\pm$ 2.194} & 46.67 {\scriptsize $\pm$ 1.847} \\
      CL-WES & 37.53 {\scriptsize $\pm$ 1.317} & 21.70 {\scriptsize $\pm$ 1.042} & 32.96 {\scriptsize $\pm$ 2.351} & 39.14 {\scriptsize $\pm$ 1.959} & 46.01 {\scriptsize $\pm$ 1.640} & 41.95 {\scriptsize $\pm$ 1.842} \\
      CL-WESS & 52.68 {\scriptsize $\pm$ 1.346} & 34.49 {\scriptsize $\pm$ 0.906} & 45.00 {\scriptsize $\pm$ 2.158} & 56.83 {\scriptsize $\pm$ 2.124} & 57.06 {\scriptsize $\pm$ 1.014} & 53.73 {\scriptsize $\pm$ 1.387} \\
      \hline
      Average fusion & 81.34 {\scriptsize $\pm$ 1.329} &	65.78 {\scriptsize $\pm$ 1.470} & 61.87 {\scriptsize $\pm$ 0.749} & 91.87 {\scriptsize $\pm$ 0.452} & 79.77 {\scriptsize $\pm$ 1.106} & 75.82 {\scriptsize $\pm$ 0.972} \\
      Weighed fusion & 84.61 {\scriptsize $\pm$ 2.873} & 69.69 {\scriptsize $\pm$ 1.660} & 67.02 {\scriptsize $\pm$ 0.935} & 94.38 {\scriptsize $\pm$ 0.502} & 83.74 {\scriptsize $\pm$ 0.490} & 80.01 {\scriptsize $\pm$ 0.623} \\
      Decision Tree & 95.25 {\scriptsize $\pm$ 1.761} & 74.10 {\scriptsize $\pm$ 1.288} & 72.19 {\scriptsize $\pm$ 1.437} & 97.05 {\scriptsize $\pm$ 1.193} & 95.16 {\scriptsize $\pm$ 1.149} & 89.15 {\scriptsize $\pm$ 1.230 } \\
      \hline
      \hline
      \multicolumn{7}{|c|}{\bf Sentence level} \\
      \hline
      \bf Methods & \bf Wikipedia (\%) & \bf TALN (\%) & \bf JRC (\%) & \bf APR (\%) & \bf Europarl (\%) & \bf Overall (\%) \\
      \hline
      CL-C3G & 48.24 {\scriptsize $\pm$ 0.272} & 48.19 {\scriptsize $\pm$ 0.520} & 36.85 {\scriptsize $\pm$ 0.727} & 61.30 {\scriptsize $\pm$ 0.567} & 52.70 {\scriptsize $\pm$ 0.928} & 49.34 {\scriptsize $\pm$ 0.864} \\
      CL-CTS & 46.71 {\scriptsize $\pm$ 0.388} & 38.93 {\scriptsize $\pm$ 0.284} & 28.38 {\scriptsize $\pm$ 0.464} & 51.43 {\scriptsize $\pm$ 0.687} & 53.35 {\scriptsize $\pm$ 0.643} & 47.50 {\scriptsize $\pm$ 0.601} \\
      CL-ASA & 27.68 {\scriptsize $\pm$ 0.336} & 27.33 {\scriptsize $\pm$ 0.306} & 34.78 {\scriptsize $\pm$ 0.455} & 25.95 {\scriptsize $\pm$ 0.604} & 36.73 {\scriptsize $\pm$ 1.249} & 35.81 {\scriptsize $\pm$ 1.036} \\
      CL-ESA & 50.89 {\scriptsize $\pm$ 0.902} & 14.41 {\scriptsize $\pm$ 0.233} & 14.45 {\scriptsize $\pm$ 0.380} & 14.18 {\scriptsize $\pm$ 0.645} & 14.09 {\scriptsize $\pm$ 0.583} & 14.44 {\scriptsize $\pm$ 0.540} \\
      T+MA & 50.39 {\scriptsize $\pm$ 0.898} & 37.66 {\scriptsize $\pm$ 0.365} & 32.31 {\scriptsize $\pm$ 0.370} & 61.95 {\scriptsize $\pm$ 0.706} & 37.70 {\scriptsize $\pm$ 0.514} & 37.42 {\scriptsize $\pm$ 0.490} \\
      \hline
      CL-CTS-WE & 47.26 {\scriptsize $\pm$ 1.647} & 43.93 {\scriptsize $\pm$ 1.881} & 31.63 {\scriptsize $\pm$ 0.904} & 57.85 {\scriptsize $\pm$ 1.921} & 56.39 {\scriptsize $\pm$ 2.032} & 50.69 {\scriptsize $\pm$ 1.767} \\
      CL-WES & 28.48 {\scriptsize $\pm$ 0.865} & 24.37 {\scriptsize $\pm$ 0.720} & 33.99 {\scriptsize $\pm$ 0.903} & 39.10 {\scriptsize $\pm$ 0.863} & 44.06 {\scriptsize $\pm$ 1.399} & 41.43 {\scriptsize $\pm$ 1.262} \\
      CL-WESS & 45.65 {\scriptsize $\pm$ 2.100} & 40.45 {\scriptsize $\pm$ 1.837} & 48.64 {\scriptsize $\pm$ 1.328} & 58.08 {\scriptsize $\pm$ 2.459} & 58.84 {\scriptsize $\pm$ 1.769} & 56.35 {\scriptsize $\pm$ 1.695} \\
      \hline
      Decision Tree & 80.45 {\scriptsize $\pm$ 1.658} & 80.89 {\scriptsize $\pm$ 0.944} & 72.70 {\scriptsize $\pm$ 1.446} & 78.91 {\scriptsize $\pm$ 1.005} & 94.04 {\scriptsize $\pm$ 1.138} & 88.50 {\scriptsize $\pm$ 1.207} \\
      \hline
\end{tabular}
\end{small}
 \end{center}
\caption{\label{table_res} Average~$F_{1}$~scores and confidence intervals of cross-language similarity detection methods applied on \mbox{EN$\rightarrow$FR} sub-corpora -- 8 folds validation.}
\end{table*}

\section{Combining multiple methods}
\label{fusion}
\subsection{Weighted Fusion}
\label{w_fusion}

We try to combine our methods to improve cross-language similarity detection performance. 
During weighted fusion, we assign one weight to the similarity score of each method and we calculate a (weighted) composite score. 
We optimize the distribution of the weights with \emph{\mbox{Condor}} \cite{condor}.
We use the first two folds of each sub-corpus to determinate the optimal weights, while the other eight folds evaluate the fusion.
We also try an average fusion, \textit{i.e.} a weighted fusion where all the weights are equal.

\subsection{Decision Tree Fusion}
\label{decision_tree}

\begin{figure}[ht]
    \centering
    \subfloat[Distribution histogram (fingerprint) of \emph{\mbox{CL-C3G}}]{{\includegraphics[width=7.5cm]{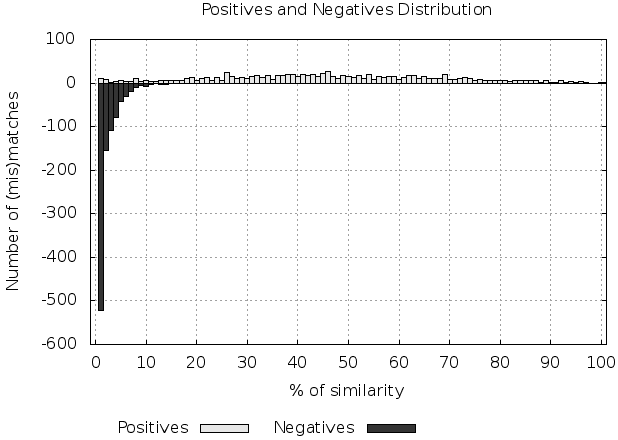} }}
    \qquad
    \subfloat[Distribution histogram (fingerprint) of \emph{\mbox{CL-ASA}}]{{\includegraphics[width=7.5cm]{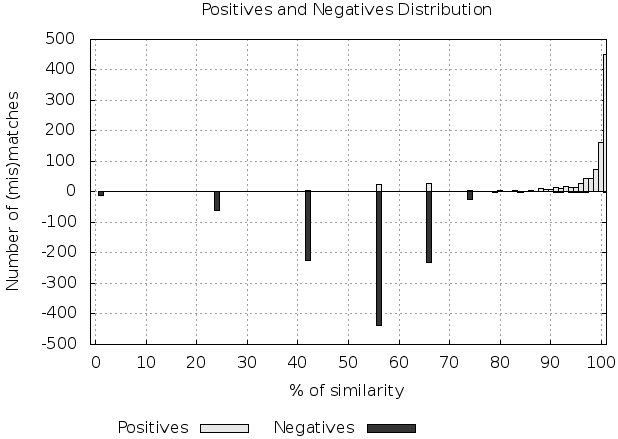} }}
    \caption{Distribution histograms of two state-of-the-art methods for 1000~positives and 1000~negatives (mis)matches.}
    \label{fingerprint}
    \vspace{-0.5cm}
\end{figure}

Regardless of their capacity to predict a (mis)match, an interesting feature of the methods is their clustering capacity, \textit{i.e.} their ability to correctly separate the positives (similar units) and the negatives (different units) in order to minimize the doubts on the classification.
Distribution histograms on Figure~\ref{fingerprint} highlight the fact that each method has its own fingerprint.
Even if two methods look equivalent in term of final performance, their distribution can be different.
One explanation is that the methods do not process on the same way.
Some methods are lexical-syntax-based, others process by aligning concepts (more semantic) and still others capture context with word vectors.
For instance, \emph{\mbox{CL-C3G}} has a narrow distribution of negatives and a broad distribution for positives (Figure~\ref{fingerprint} (a)), whereas the opposite is true for \emph{\mbox{CL-ASA}} (Figure~\ref{fingerprint} (b)).
We try to exploit this complementarity using decision tree based fusion.
We use the C4.5 algorithm \cite{quinlan1993} implemented in \emph{\mbox{Weka}}~3.8.0 \cite{weka}.
The first two folds of each sub-corpus are used to determinate the optimal decision tree and the other eight folds to evaluate the fusion (same protocol as weighted fusion).
While analyzing the trained decision tree, we see that \emph{\mbox{CL-C3G}}, \emph{\mbox{CL-WESS}} and \emph{\mbox{CL-CTS-WE}} are the closest to the root. 
This confirms their relevance for similarity detection, as well as their complementarity.

\section{Results and Discussion}
\label{result}

\textbf{Use of word embeddings.} We can see in Table~\ref{table_res} that the use of distributed representation of words instead of lexical resources improves \emph{\mbox{CL-CTS}} (\emph{\mbox{CL-CTS-WE}} obtains overall performance gain of +3.83\% on chunks and +3.19\% on sentences).
Despite this improvement, CL-CTS-WE remains less efficient than \emph{\mbox{CL-C3G}}.
While the use of bilingual sentence vector (\emph{\mbox{CL-WES}}) is simple and elegant, its performance is lower than three state-of-the-art methods. 
However, its syntactically weighted version (\emph{\mbox{CL-WESS}}) looks very promising and boosts the \emph{\mbox{CL-WES}} overall performance by +11.78\% on chunks and +14.92\% on sentences.
Thanks to this improvement, \emph{\mbox{CL-WESS}} is significantly better than \emph{\mbox{CL-C3G}} (+2.97\% on chunks and +7.01\% on sentences) and is the best single method evaluated so far on our corpus.

\textbf{Fusion.} Results of the decision tree fusion are reported at both chunk and sentence level in Table~\ref{table_res}. 
Weighted and average fusion are only reported at chunk level.
In each case, we combine the 8 previously presented methods (the 5 state-of-the-art and the 3 new methods).
Weighted fusion outperforms the state-of-the-art and the embedding-based methods in any case.
Nevertheless, fusion based on a decision tree looks much more efficient. 
At chunk level, decision tree fusion leads to an overall $F_{1}$~score of 89.15\% while the precedent best weighted fusion obtains 80.01\% and the best single method only obtains 53.73\%. 
The trend is the same at the sentence level where decision tree fusion largely overpasses any other method (88.50\% against 56.35\% for the best single method).
%In our evaluation, best minimal decision tree involves \mbox{\textit{CL-C3G}}, \mbox{\textit{CL-WESS}} and \mbox{\textit{CL-CTS-WE}}, for an overall higher than 85\% of correct classification on both levels.
In our evaluation, the best decision tree, for an overall higher than 85\% of correct classification on both levels, involves at a minimum \mbox{\textit{CL-C3G}}, \mbox{\textit{CL-WESS}} and \mbox{\textit{CL-CTS-WE}}.
These results confirm that different methods proposed complement each other, and that embeddings are useful for cross-language textual similarity detection.

\section{Conclusion and Perspectives}
\label{conclusion}

We have augmented several baseline approaches using word embeddings. 
The most promising approach is a cosine similarity on syntactically weighted distributed representation of sentence (\emph{\mbox{CL-WESS}}), which beats in overall the precedent best state-of-the-art method.
Finally, we have also demonstrated that all methods are complementary and their fusion significantly helps cross-language textual similarity detection performance.
At chunk level, decision tree fusion leads to an overall $F_{1}$~score of 89.15\% while the precedent best weighted fusion obtains 80.01\% and the best single method only obtains 53.73\%. 
The trend is the same at the sentence level where decision tree fusion largely overpasses any other method.

Our future short term goal is to work on the improvement of \emph{\mbox{CL-WESS}} by analyzing the syntactic weights or even adapt them according to the plagiarist's stylometry. 
We have also made a submission at the SemEval-2017 Task~1, \textit{i.e.} the task on Semantic Textual Similarity detection.

\bibliography{eacl2017}
\bibliographystyle{eacl2017}

\appendix

\end{document}